\documentclass[sn-basic,iicol]{sn-jnl}

\usepackage{amsmath,amsfonts,bm}

\def\eqref#1{equation~\ref{#1}}

\def\1{\bm{1}}

\def\ve{{\bm{e}}}

\def\vk{{\bm{k}}}

\def\vm{{\bm{m}}}

\def\vq{{\bm{q}}}
\def\vr{{\bm{r}}}

\def\vv{{\bm{v}}}
\def\vw{{\bm{w}}}
\def\vx{{\bm{x}}}
\def\vy{{\bm{y}}}

\def\mR{{\bm{R}}}

\DeclareMathAlphabet{\mathsfit}{\encodingdefault}{\sfdefault}{m}{sl}
\SetMathAlphabet{\mathsfit}{bold}{\encodingdefault}{\sfdefault}{bx}{n}

\usepackage{graphicx}%
\usepackage{multirow}%
\usepackage{amsmath,amssymb,amsfonts}%
\usepackage{amsthm}%
\usepackage{mathrsfs}%
\usepackage[title]{appendix}%
\usepackage{xcolor}%
\usepackage{textcomp}%
\usepackage{manyfoot}%
\usepackage{booktabs}%
\usepackage{algorithm}%
\usepackage{algorithmicx}%
\usepackage{algpseudocode}%
\usepackage{listings}%
\usepackage{comment}

\usepackage{makecell}

\newcommand{\CS}{{\textsuperscript{13}C}}
\newcommand{\HS}{{\textsuperscript{1}H}}

\newcommand{\Name}{{GeqShift}}

\newcommand{\DHS}{$\delta_\text{H}$}
\newcommand{\DCS}{$\delta_\text{C}$}
\usepackage{lineno}

\begin{document}

\title[Article Title]{Carbohydrate NMR chemical shift predictions using E(3) equivariant graph neural networks}



\author*[1,2]{\fnm{Maria} \sur{Bånkestad}}\email{maria.bankestad@it.uu.se}
  
\author[3]{\fnm{Kevin M.} \sur{Dorst}}\email{kevin.dorst@su.se}
  
\author[3]{\fnm{Göran} \sur{Widmalm}}\email{goran.widmalm@su.se}
\author[1]{\fnm{Jerk} \sur{Rönnols}}\email{jerk.ronnols@ri.se}

\affil*[1]{\orgdiv{RISE Research Institutes of Sweden}, \country{Stockholm}}

\affil[2]{\orgdiv{Department of Information Technology}, \orgname{Uppsala University}, \orgaddress{ \country{Sweden}}}

\affil[3]{\orgdiv{Department of Organic Chemistry}, \orgname{Stockholm University}, \orgaddress{  \country{Sweden}}}


\abstract{Carbohydrates, vital components of biological systems, are well-known for their structural diversity. Nuclear Magnetic Resonance (NMR) spectroscopy plays a crucial role in understanding their intricate molecular arrangements and is essential in assessing and verifying the molecular structure of organic molecules. An important part of this process is to predict the NMR chemical shift from the molecular structure. This work introduces a novel approach that leverages E(3) equivariant graph neural networks to predict carbohydrate NMR spectra. Notably, our model achieves a substantial reduction in mean absolute error, up to threefold, compared to traditional models that rely solely on two-dimensional molecular structure.
 Even with limited data, the model excels, highlighting its robustness and generalization capabilities. The implications are far-reaching and go beyond an advanced understanding of carbohydrate structures and spectral interpretation. For example, it could accelerate research in pharmaceutical applications, biochemistry, and structural biology, offering a faster and more reliable analysis of molecular structures. Furthermore, our approach is a key step towards a new data-driven era in spectroscopy, potentially influencing spectroscopic techniques beyond NMR.}




\maketitle

\section{Introduction}\label{sec:intro}

Carbohydrates are intricate organic compounds that ubiquitously occur in all living organisms. Their significance spans across all domains of life, but especially in cell-cell interactions and disease processes. In recent decades, a remarkable advancement in our comprehension of carbohydrate chemistry and biology has been attributed to their vital importance. The molecular structure of carbohydrates is notably complex and diverse and, therefore, challenging for chemists to construct and manipulate \citep{peterson2021rapid,dal2023linker}. The role of carbohydrates in biological processes heavily depends on their three-dimensional structures, which include the covalent bonds and the conformations these molecules adopt over time. Nuclear magnetic resonance (NMR) spectroscopy is a fundamental technique to decipher the intricate three-dimensional structure of molecules. This study introduces a cutting-edge machine-learning model to interpret NMR spectra, which considers molecule geometries and known symmetries.\looseness=-1

The inherent complexity of carbohydrate molecules in structural studies and stereochemical assignments stems from two key factors: their large number of stereocenters and the extensive possibilities for interconnecting monosaccharides. For example, combining two glucopyranosyl residues can yield as many as 19 distinct disaccharides, each with a unique structure \citep{Roslund2008}. Additionally, variations in substitution patterns, like acetylation and sulfonation, further contribute to the complexity of carbohydrate structures. Determining carbohydrate structures by NMR spectroscopy can be a formidable task \citep{fontana2023primary}.

\begin{figure}[t]
     \begin{center}
    \includegraphics[width=0.98\linewidth]{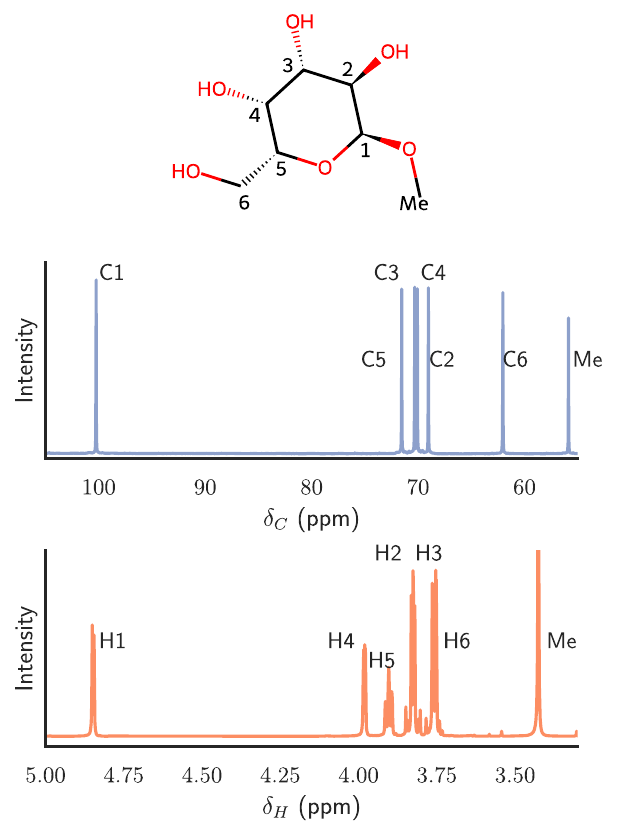}
  \end{center}
    \vspace{0.2cm}
  \caption{The {\CS} and {\HS} NMR spectrum of \mbox{methyl $\alpha$-\textsc{d}-galactopyranoside}. The peaks of the specific carbons (from C1 to C6 and the \textit{O}-methyl group) with their connected hydrogens are indicated in the plot.}
  \label{fig:spectrums_h_c_aDGalOme}
\end{figure}

The peaks observed in an NMR spectrum of a molecule provide valuable information about the presence of nuclei  and their chemical surroundings, such as carbon and hydrogen isotopes \textsuperscript{13}C and \textsuperscript{1}H, and how they are interconnected. Figure \ref{fig:spectrums_h_c_aDGalOme} provides examples of {\CS} and {\HS} NMR spectra for a monosaccharide.

\begin{figure*}[th]
    \centering
    \includegraphics[width = 0.85\linewidth]{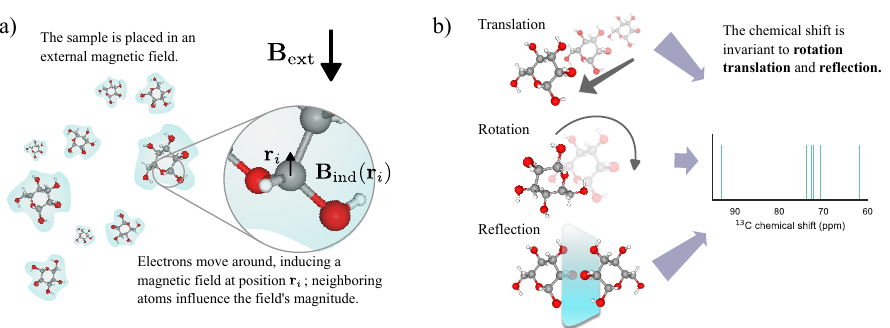}
    \vspace{0.3cm}
    \caption{\textbf{a}: The compound under examination moves within a fluid environment and interacts with an external magnetic field denoted as $\mathbf{B}_{\text{ext}}$. An induced magnetic field $\mathbf{B}_{\text{ind}}(\mathbf{r}_i)$ at a specific position $\mathbf{r}_i$ determines the chemical shift of a resonating nucleus. \textbf{b}: The chemical shift $\delta$ remains constant under the Euclidean group E(3), i.e., it is unaffected by translation, rotation, and reflection.}
    \label{fig:invariant}
\end{figure*}

The position of a peak for a particular nucleus, indicated by its chemical shift $\delta$ ({\DHS} and {\DCS} for {\HS} and {\CS} chemical shifts, respectively), corresponds to the resonance frequency of the nucleus within a magnetic field. The local environment of the atom, especially the electron density in the vicinity of the nucleus, strongly influences this resonance frequency (see Figure \ref{fig:invariant}). Besides the atomic species of the studied nucleus, the primary factors influencing chemical shifts are the neighboring covalently bonded atoms within the molecule because the electronegativity of these nearby atoms correlates closely with the observed chemical shifts. Electron-withdrawing groups, like oxygen and fluorine, located near the observed nuclei deshield them, increasing their chemical shifts. Conversely, proximity to electron-donating groups enhances shielding, decreasing the chemical shifts.\looseness=-1

\begin{figure}[ht]
     \begin{center}
    \includegraphics[width=0.63\linewidth]{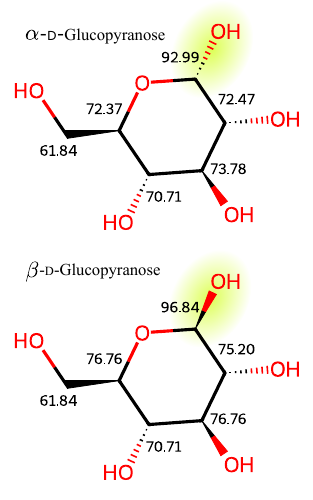}
  \end{center}
  \vspace{0.2cm}
  \caption{\textsuperscript{13}C NMR chemical shifts of two glucose isomers, $\alpha$-\textsc{d}-glucopyranose and $\beta$-\textsc{d}-glucopyranose. These isomers differ only in the stereochemistry of the anomeric center (highlighted). This subtle variation substantially impacts the chemical shifts in an NMR spectrum. }
  \label{fig:glucopyranose}
\end{figure}

In molecular ring systems (appearing in carbohydrates), the orientation of a hydrogen atom, either axially or equatorially, significantly impacts its $\delta_H$ value. Similarly, for carbon nuclei in a ring system, the arrangement of substituents they carry influences their $\delta_C$ value. Figure \ref{fig:glucopyranose}, showing the {\CS} chemical shifts of $\alpha$- and $\beta$-glucopyranose, illustrates this discrepancy. The change in configuration at the anomeric center not only affects the highlighted anomeric carbon but also has a ripple effect, altering the shifts of all carbon atoms in the molecule. It is important to note that spatial interactions can influence chemical shifts beyond the effects of covalent bonds \citep{Kwon2023_nonconventionalHBond}.

 A standard method for predicting the chemical shifts of carbohydrate molecules involves utilizing an extensive database of known carbohydrates \citep{Loss2015}. This approach entails comparing new carbohydrate structures with those existing in the database, making necessary adjustments for recognized patterns around glycosidic bonds. 
 
 The CASPER program \citep{lundborg2011structural} relies on a relatively small set of NMR data of glycans and uses approximations to predict chemical shifts of glycan structures not present in the database, which facilitates the coverage of a large number of structures. However, the reliance on these databases is less effective when new structures containing previously uncharacterized sugar residues are encountered.  

Alternatively, chemical shifts can be estimated using Quantum Mechanical Density Functional Theory (DFT) calculations \citep{argaman2000density}. While this technique is effective for many molecules, it comes with substantial computational demands, making it both costly and time-consuming. A notable advancement in carbohydrate chemical shift calculation was recently published by \citep{Palivec2022} and involves an in-depth simulation of the water environment surrounding the molecules under study. This approach employs molecular dynamics and DFT to calculate chemical shifts for small carbohydrate molecules, including mono-, di-, and one trisaccharide. The results from the study by \citep{Palivec2022} serve as a basis for comparing our findings.

\begin{figure*}[t]
    \centering
    \includegraphics[width = 0.87\linewidth]{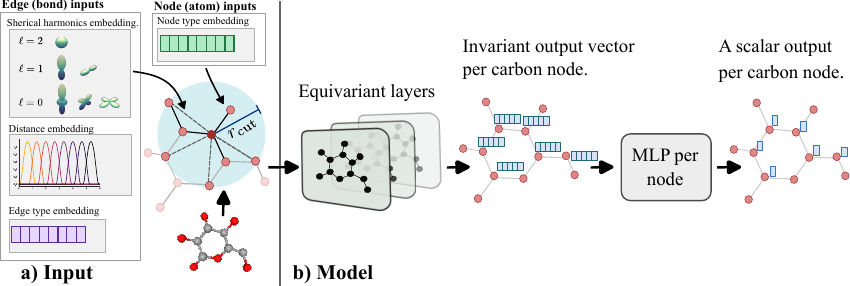}
    \vspace{0.5cm}
    \caption{An overview of the model. The left side (labeled \textbf{a}) shows the components involved in processing molecule input data, including node embeddings with atom type and neighboring hydrogen information and edge embeddings representing bond types and relative distances between connected nodes. The $r_{cut}$ parameter denotes the cutoff radius for defining neighboring atoms. The model architecture is illustrated on the right side (labeled \textbf{b}). It consists of $K$ equivariant layers, with the final layer producing an invariant vector for each node. Nodes containing chemical shift data are processed individually, passing through a multi-layer perceptron (MLP) to generate an invariant chemical shift prediction. }
    \label{fig:model_overview}
\end{figure*}

As mentioned earlier, the relationship between a molecule and its chemical shift is intricate, which makes one consider using artificial neural networks, known as universal approximators, to capture this relationship from data. Already 1991, \citet{doi:10.1126/science.1990429} suggested using a feed-forward network to identify $^1$H NMR spectra for oligosaccharides. In recent years, graph neural networks (GNN) have emerged to learn chemical shifts \citep{jonas2022prediction}. Some of these models rely solely on the molecular structure (the atoms and their bonds) as their input  \citep{kwon2020neural,yang2021predicting,han2022scalable}. In contrast, others have introduced atom-atom pairwise distances as additional input features \citep{gerrard2020impression, guan2021real}.

While these models demonstrate strong performance for numerous molecules, they struggle when dealing with molecules featuring complex stereochemistry, such as carbohydrates. It is appropriate to assume that these molecules must be treated as dynamic, three-dimensional entities for accurate representation, demanding a network capable of capturing this complexity. This study proposes a model that integrates the three-dimensional molecular structure while preserving the fundamental symmetries of the molecule's underlying physics. \looseness=-1


More specifically, we introduce an E(3) equivariant graph neural network, also known as an Euclidean neural network \citep{geiger2022e3nn}. Equivariance is a transformation property that assures a consistent response when a feature transforms. An example of equivariance is the intramolecular forces holding the atoms together in a molecule. These forces are equivariant to rotation since these forces rotate together with the molecule. During transformations, an equivariant function preserves relationships between input ( molecule) and output (interatomic forces). If we have an equivariant function deriving the interatomic forces, these derived forces rotate with the molecule.\looseness=-1

An Euclidean neural network is equivariant to the Euclidean group E(3), which is the group of transformations in the Euclidean space, including rotation, translation, and mirroring. Compared to a network that solely considers pairwise distance, an equivariant network considers the relative distance between atoms, encompassing both pairwise distance and pairwise direction. Euclidean neural networks have recently gained recognition for their success in various chemistry applications, spanning from modeling molecule potential energy surfaces \citep{batzner20223} to predicting toxicity \citep{cremer2023equivariant} and studying protein folding \citep{jumper2021highly}.

Our model, denoted as \emph{{\Name}} (geometric equivariant shift), is a GNN that utilizes equivariant graph self-attention layers \citep{liao2023equiformer} to learn chemical shifts, particularly when stereochemistry is crucial. These attention layers update the node features by considering features of close nodes, so-called neighbors, and weights these neighbors to emphasize the most important information, using so-called attention weights. Our contribution is three-fold: the chemical shift prediction model {\Name}, an innovative data augmentation method inspired by the dynamic movement of molecules in a fluid, and a compiled carbohydrate chemical shift dataset suitable for machine learning applications. By making this dataset public, we hope to stimulate further research in data-driven automated chemical shift analysis. 

Ablation studies demonstrate that our model and training approach achieve precise predictions, especially in intricate stereochemistry cases. Notably, for the carbohydrate dataset, our network reaches mean absolute errors (MAEs) of 0.31 for {\DCS} and 0.032 for {\DHS}. These MAE values are over three times lower for {\DCS} and two times lower for {\DHS} compared to models relying solely on two-dimensional molecular structure information.\looseness=-1

\section{Results}\label{sec:results}
Our model is trained on {\CS} and {\HS} NMR chemical shift data from the CASPER program \citep{lundborg2011structural}, which is further detailed in the methods section. The performance is assessed using a ten-fold cross-validation approach, where each split maintains a balanced distribution of mono-, di-, and trisaccharides. Each split comprises approximately 336 carbohydrate structures for training and 39 for testing.

A molecule is inherently dynamic, continuously changing its conformation. The likelihood of these conformations follows the Boltzmann distribution, $p(\mathbf{R}) \sim \exp(-E(\mathbf{R}))$, where $E$ is the molecule energy function and $\mathbf{R}$ its conformation. 
Conventionally, in data-driven models, this problem is alleviated by selecting the conformation with the lowest energy, implying the highest probability. This is typically determined through methods like density function theory (DFT) simulation.\looseness-1

We take a different approach by considering the molecule conformation as dynamic, with not just one but an ensemble of conformations. The predicted shift varies depending on the conformation, resulting in an ensemble of predictions per molecule. The final prediction is the ensemble average. We use this technique during both training and testing.

In machine learning terms, this is a data augmentation technique. We hypothesize that this will enhance the generalization capacity of the model, especially given the limited size of the training dataset. As a result, our final model, {\Name}, does not rely on a specific low-energy conformation as input, enabling effective generalization to molecules not seen during training. Figure \ref{fig:model_overview} presents an overview of the model. \looseness=-1

To establish a baseline, we compare our model with the scalable GNN introduced in \citep{han2022scalable}, referred to as \mbox{SG-IMP-IR}. Additionally, we conducted six ablations to assess the effectiveness of various components in our model, as summarized in Table \ref{tab:model_recap}. These evaluations include comparing the use of an invariant version (inv) of the model, the same as setting $\ell_{max} = 0$, the maximum degree of the irreducible representations of the hidden layers (explained further in Section \ref{sec:background}). Furthermore, we examined the impact of testing and training on an ensemble of conformations by evaluating the model on only a single conformation (1T) and training and testing on a single conformation (1TT). It is important to note that the train/test splits are consistent across all models, with data augmentation achieved by sampling multiple conformations per molecule. \looseness=-1

\begin{table}[ht]
\vspace{-0.3cm}
\setlength{\tabcolsep}{2pt}
\caption{An overview of our two models with their training and test data variations.}\label{tab:model_recap}%
\begin{tabular}{@{}lccc@{}}
\toprule
Models &  \thead{Nbr conf. \\ per mol \\ in train} &  \thead{Nbr conf. \\ per mol \\  in test}& \thead{$\ell_{max}$ in \\ hidden layers} \ \\
\midrule
{\Name}\_1TT\_inv   & 1  & 1  &0  \\
{\Name}\_1TT    & 1   & 1  & 2  \\
{\Name}\_1T\_inv    & 100   & 1  & 0 \\
{\Name}\_1T   & 100   & 1  & 2  \\
{\Name}\_inv    & 100   & 100  & 0  \\
{\Name}   & 100   & 100  & 2  \\
\botrule
\end{tabular}
\end{table}

\begin{figure}[t]
    \centering
\includegraphics[width = 0.999\linewidth]{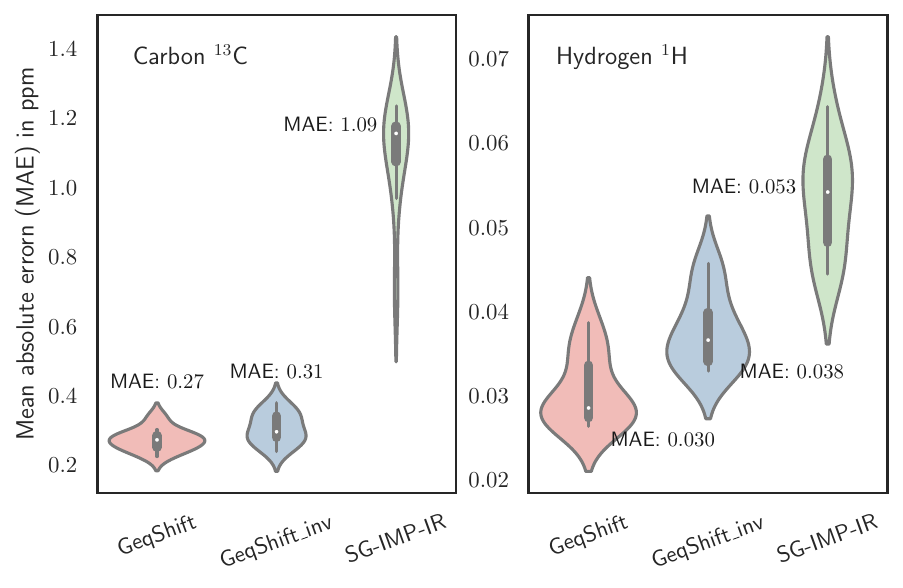}
    \caption{Comparison of the test prediction accuracy in mean absolute error MAE between the baseline model \mbox{SG-IMP-IR} and our proposed model {\Name}, and its invariant version {\Name}\_inv. The result is visualized using violin plots.}    
    \label{fig:tot_results_all_models}
\end{figure}

Figure \ref{fig:tot_results_all_models} presents an overview of the model's performance using violin plots, a combination of a box plot, and a density plot \citep{hintze1998violin}. Furthermore, Table \ref{tab:res_carbon} provides a detailed comparison of the models, emphasizing prediction accuracy for different types of carbohydrates, including mono-, di-, and trisaccharides.\looseness=-1

Among our models, {\Name} emerges as the top-performing model, closely followed by {\Name}\_inv. Compared to using just one conformation per molecule for training, we observe a significant performance improvement when using an ensemble of 100 conformations. For instance, in the case of monosaccharides, the mean absolute error (MAE) notably decreases from 0.55 to 0.37 when trained with 100 conformations. Subsequently, it further drops slightly to 0.31 when also predicting 100 conformations. These results underscore the advantage of incorporating multiple conformations in our training and prediction processes.\looseness=-1

Comparing the MAE values of {\CS} at 0.27 and {\HS} at 0.030 with the best-performing DFT-calculated chemical shifts in \citet{Palivec2022}, which reported an MAE of 1.11 for {\CS} and 0.081 for {\HS}, demonstrates our model's superior predictive accuracy.

\begin{table*}[ht]
\setlength{\tabcolsep}{5pt}
\small
\caption{Comparison of prediction test accuracy for $^{13}$C and $^{1}$H chemical shifts in terms of MAE (ppm) and RMSE (ppm) split between monosaccharides, disaccharides, and trisaccharides. The accuracy is presented as the ten-fold mean, standard deviation in parenthesis. 
} \label{tab:res_carbon}%
\begin{tabular*}{\linewidth}{lcccccc}
\toprule 
 & \multicolumn{2}{c}{Monosaccharides $^{13}$C} & \multicolumn{2}{c}{Disaccharides $^{13}$C} & \multicolumn{2}{c}{Trisaccharides $^{13}$C} \\
 & MAE & RMSE & MAE & RMSE & MAE & RMSE \\
\midrule
SG-IMP-IR & 1.18 (0.20) & 1.61 (0.30) & 1.02 (0.17) & 1.53 (0.37) & 1.13 (0.16) & 1.61 (0.20) \\
{\Name}\_1TT\_inv & 0.54 (0.12) & 0.86 (0.23) & 0.44 (0.07) & 0.73 (0.15) & 0.65 (0.11) & 1.06 (0.21) \\
{\Name}\_1TT & 0.55 (0.15) & 0.90 (0.37) & 0.47 (0.08) & 0.75 (0.17) & 0.63 (0.11) & 1.05 (0.24) \\
{\Name}\_1T\_inv & 0.39 (0.11) & 0.69 (0.23) & 0.28 (0.06) & 0.51 (0.16) & 0.37 (0.10) & 0.64 (0.21) \\
{\Name}\_1T & 0.34 (0.08) & 0.61 (0.19) & 0.25 (0.06) & 0.48 (0.18) & 0.33 (0.09) & 0.57 (0.20) \\
{\Name}\_inv  & 0.37 (0.11) & 0.66 (0.23) & 0.26 (0.06) & 0.49 (0.16) & 0.33 (0.08) & 0.59 (0.14) \\
{\Name}  & \textbf{0.31} (0.08) & \textbf{0.58} (0.18) & \textbf{0.23} (0.06) & \textbf{0.46} (0.19) & \textbf{0.30} (0.09) & \textbf{0.53} (0.16) \\

\toprule
 & \multicolumn{2}{c}{Monosaccharides $^{1}$H} & \multicolumn{2}{c}{Disaccharides $^{1}$H} & \multicolumn{2}{c}{Trisaccharides $^{1}$H} \\
 & MAE & RMSE & MAE & RMSE & MAE & RMSE \\
\midrule
SG-IMP-IR & 0.071(0.026) & 0.110(0.039) & 0.045(0.007) & 0.075(0.014) & 0.055(0.011) & 0.087(0.020) \\
{\Name}\_1TT\_inv & 0.064(0.011) & 0.100(0.022) & 0.049(0.009) & 0.076(0.017) & 0.067(0.009) & 0.102(0.015) \\
{\Name}\_1TT & 0.061(0.016) & 0.115(0.053) & 0.041(0.006) & 0.061(0.012) & 0.060(0.010) & 0.103(0.030) \\
{\Name}\_1T\_inv  & 0.046(0.014) & 0.078(0.040) & 0.034(0.006) & 0.053(0.012) & 0.050(0.010) & 0.079 (0.018) \\
{\Name}\_1T  & 0.037(0.009) & 0.062(0.020) & 0.028(0.003) & 0.046(0.010) & 0.038(0.009) & 0.057 (0.017) \\
{\Name}\_inv  & 0.044(0.015) & 0.077(0.041) & 0.030(0.004) & 0.048(0.011) & 0.043(0.009) & 0.069(0.015) \\
{\Name}  & \textbf{0.035}(0.009) & \textbf{0.057}(0.018) & \textbf{0.026}(0.003) & \textbf{0.044}(0.011) & \textbf{0.033}(0.009) & \textbf{0.052}(0.016) \\
\bottomrule
\end{tabular*}
\end{table*}

\begin{figure}[ht]
    \centering
\includegraphics[width = 0.99\linewidth]{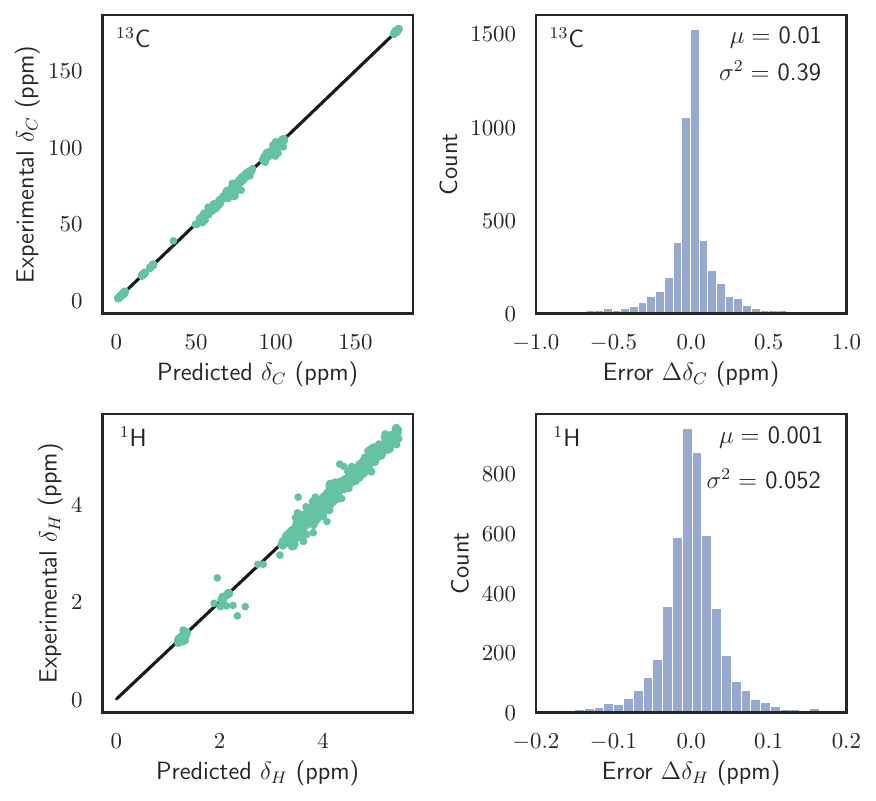}
    \caption{The figure examines the test prediction outcomes of our proposed method, {\Name}. To the left, scatter plots illustrate the relationship between actual and predicted values. Histograms representing the distribution of prediction errors $\Delta \delta$ are shown on the right.}
    \label{fig:all_carbohydrates_best_model}
\end{figure}

In Figure \ref{fig:all_carbohydrates_best_model}, we delve deeper into the prediction accuracy of our best-performing method, {\Name}. The figures within this plot illustrate histograms of prediction errors and scatter plots depicting the relationship between the actual and predicted values for both {\CS} and {\HS} nuclei. We combined the test sets' prediction results across all ten cross-validation folds to create these visualizations. Notably, the distributions of prediction errors are approximately zero-centered, with a standard deviation of $0.39$ for {\CS} and $0.052$ for {\HS}.\looseness=-1

\begin{figure*}[ht]
\centering
    \includegraphics[width = 0.7\linewidth]{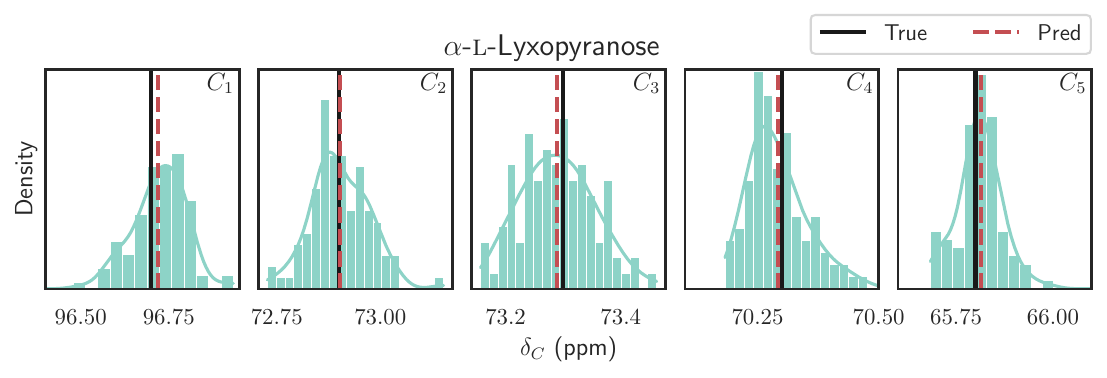}
 \caption{A histogram representing the test predictions of {\CS} chemical shifts obtained from 100 different molecular geometries of the monosaccharide $\alpha$-\textsc{l}-lyxopyranose. We highlight the prediction mean and the actual peak value. While various geometries yield slightly different chemical shift values, the average of these peaks closely approximates the experimentally determined value.} \label{fig:aLLyxose_dist}
\end{figure*}
Figure \ref{fig:aLLyxose_dist} visualizes the predictions from the whole ensemble of conformations for the monosaccharide
 $\alpha$-\textsc{l}-lyxopyranose. The figure displays histograms representing the predictions for each {\CS} atom in the molecule, the ensemble mean, and the actual NMR peaks. These histograms showcase the distribution of predicted values, allowing for a comparison with a real NMR spectrum (refer to Figure \ref{fig:spectrums_h_c_aDGalOme}). 
Furthermore, the ensemble of predictions per chemical shift enables an estimation of prediction uncertainty by examining the standard deviation.

\subsubsection*{Polysaccharides}
\begin{figure}[ht]
\centering
\includegraphics[width = 0.95\linewidth]{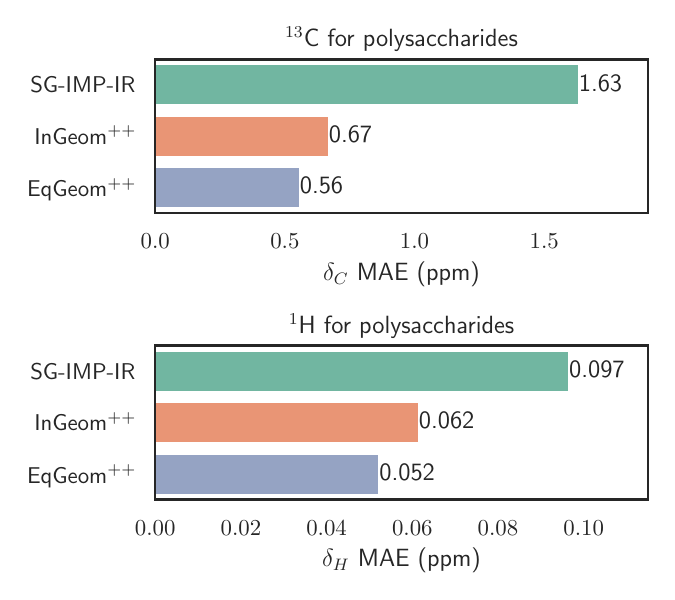}
\caption{Prediction performance for chemical shifts ({\CS} and {\HS}) in terms of mean absolute error (MAE) within the context of the two polysaccharides introduced in Figure \ref{fig:poly_errors_per_atom}. In this evaluation, the models employ an average prediction derived from the ten models trained during ten-fold cross-validation.}
\label{fig:poly_mae}
\end{figure}

\begin{figure}[ht]
    \centering
\includegraphics[width = 0.9\linewidth]{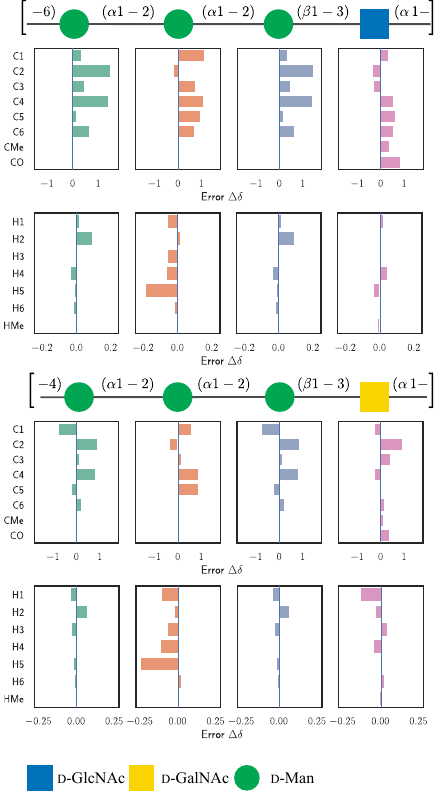}
\vspace{0.5cm}
    \caption{The figure illustrates the prediction errors for the {\CS} and {\HS} chemical shifts of two \textit{E. coli} O-antigen polysaccharides, each composed of tetrasaccharide repeating units.\citep{YILDIRIM2001179,OLSSON2008805} The structures are visualized using symbols from the SNFG standard \citep{neelamegham2019updates}. The repeating units are enclosed in square brackets. The box plots visually represent the prediction errors $\Delta \delta$ per-atom basis.}
    \label{fig:poly_errors_per_atom}
\end{figure}

In addition to predicting the mono-, di- and trisaccharides in the original dataset, we examine {\Name}'s capability to extend to larger carbohydrate structures. We predict the chemical shifts of two polysaccharides, each constructed of tetrasaccharide repeating units. In Figure \ref{fig:poly_mae}, the prediction accuracy of {\Name} is compared to {\Name}\_inv and SG-IMP-IR. Notably, {\Name} outperforms these models regarding both {\CS} and {\HS} prediction accuracy. Furthermore, Figure \ref{fig:poly_errors_per_atom} details the prediction errors using bar plots for individual {\CS} and {\HS} nuclei.

\section{Discussion}\label{sec:discussion}
This work introduces a novel machine learning model to predict chemical shifts, explicitly addressing the molecule's stereochemistry. We employed an Euclidean graph neural network that utilizes molecular structure and geometry as input to construct a model capable of capturing changes in molecule geometry in response to stereochemical alterations.

To achieve high accuracy, data augmentation that mimics the true nature of the molecules was utilized, where each molecule does not stay in one conformation but instead moves around. So, instead of one conformation, an ensemble of conformations for each molecule in the training and testing set was used.

The obtained prediction errors exceeded our expectations. It must be emphasized that the ranges of chemical shifts are approximately 0-200 for {\CS} and 0-10 for {\HS}, so the achieved prediction errors approach the levels that typically qualify as error margins in measurements. 
To further put the results into perspective, it is fitting to compare the prediction errors to other works using similar techniques for different classes of compounds and to alternative ways of calculating chemical shifts. The main results are those detailed in Table \ref{tab:res_carbon}, where our model is compared to a state-of-the-art neural network for chemical shift prediction, which has been retrained on our dataset. However, there are other sources. The prediction errors using DFT calculations obtained by \cite{Palivec2022} were 1.11 ppm for {\CS} and 0.081 ppm for {\HS}, to be compared with 0.27 ppm and 0.030 ppm using our model. 
The review by \citep{jonas2022prediction} contains a summary of MAEs obtained in efforts to predict chemical shifts up until 2021. Neither of these reaches MAEs $<$1 ppm for {\CS} or $<$0.1 ppm for {\HS}. The study by \cite{Unzueta2021Predicting} arrives at RMSEs of 0.7 ppm for {\CS}  and 0.11 ppm for {\HS}, to be compared with the highest RMSEs of 0.58 ppm for {\CS}  and 0.057 ppm for {\HS} of monosaccharides.

The developed model has great potential for predicting chemical shifts for other organic molecules, particularly compounds with asymmetric centers. This includes, among many different classes, pharmacological compounds and proteins.

Furthermore, the ability of the model to accurately predict physical observables, i.e., the chemical shifts based on the molecular structure, highly encourages future application of similar methodology for other analytical techniques, e.g., X-ray photoelectron spectroscopy and X-ray absorption spectroscopy and potentially for predicting other physical parameters. 

Most, if not all, studies of prediction methods for chemical shifts are focused on predicting chemical shifts from molecular structure. The inverse problem, where a molecular structure is generated from chemical shifts, is more compelling for experimental practice. At the same time, it is more complex. However, making proper chemical shift predictions builds a solid ground for attacking the inverse problem.

\section{Method}\label{sec:method}
In this section, we detail the model and the dataset by giving relevant background information, then explaining {\Name} in more detail, and finally describing the carbohydrate dataset.

\subsection{Background}\label{sec:background}
\subsubsection*{Graph neural network}
A graph $\mathcal{G} = (\mathcal{V}, \mathcal{E})$ consists of nodes $i \in \mathcal{V}$ and edges $i,j \in \mathcal{E}$, defining the relationships between the nodes $i$ and $j$. One can represent a molecule as a graph with the atoms as nodes and bonds as edges. To expand this to an even richer representation of the molecule, one can include additional edges between atoms close to each other in space; typically, we define a cutoff radius $r_{\text{cut}}$ and introduce edges between any two atoms that are less than the cutoff distance apart. A graph neural network consists of multiple message-passing layers. Given a node feature $\vx_i^k$ at node $i$ and edge features $\ve_{ij}^k$ between node $i$ and its neighbors $\mathcal{N}(i)$, the message passing procedure at layer $k$ is defined as
\begin{align}
    \vm_{ij}^k &= f^m(\vx_i^k, \vx_j^k, \ve_{ij}^k), \\
    \hat{\vx}_i^{k+1} &= f^a_{j \in \mathcal{N}(i)} (\vm_{ij}^k),\\
    {\vx}_i^{k+1} &= f^u(\vx_i^k,\hat{\vx}_i^{k+1}),
\end{align}
where $f^m$ is the message function, deriving the message from node $j$ to node $i$, and $ f^a_{j \in \mathcal{N}(i)}$ is the aggregating function, which aggregates all messages coming from the neighbors of node $i$, defined by $\mathcal{N}(i)$. The aggregation function is commonly just a simple summation or average. Finally, $f^u$ is the update function that updates the features for each node. A graph neural network (GNN) consists of message-passing layers stacked onto each other, where the node output from one layer is the input of the successive layer.

\subsubsection*{Equivariant convolutions }
Equivariance is a fundamental concept that appears throughout the natural world, governing the symmetry and behavior of physical systems, from subatomic particles to the organization of molecules in biological systems. It underpins the consistency and invariance of natural phenomena under various transformations, making it a crucial concept in the natural sciences.

Equivariance is an essential factor when considering NMR chemical shifts. In this study, we focus on predicting the isotropic part of the chemical shift tensor, denoted as $\delta_{\text{iso}}$, which is a scalar and remains unchanged under the Euclidean group E(3) (the group of rotation, translation, and mirroring) with respect to the input locations of the atoms. However, the actual chemical shift tensor, $\boldsymbol{\delta}$, is a second-rank tensor with an antisymmetric nature ($\ell = 2$ with even parity). While it is possible to predict the complete chemical shift tensors, as demonstrated by \citet{venetos2023machine}, molecules in solution in a laboratory setting move around relative to the external magnetic field. Consequently, it is the isotropic part of the chemical shift tensor observed in an NMR spectrum.
Even though the isotropic chemical shift is a scalar quantity, the relationships governing it are intricate. Therefore, using a model capable of accurately capturing these relationships would be advantageous.

Euclidean neural networks can represent a comprehensive set of tensor properties and operations that obey the same symmetries as symmetries of molecules. Formally, a function $f: X\to Y$ is equivariant to a group of transformations $G$ if for any input $x \in X$ and output $y\in Y$ and group element $g\in G$ that is well-defined in both $X$ and $Y$, we have that $fD_X(g)(x) = D_Y(g)f(x)$, where $D_X(g)$ and $D_Y(g)$ are transformation matrices parameterized by $g$ in $X$ and $Y$. In other words, the result is the same regardless of whether the transformation is applied before the function or vice versa. An example is if you have a function deriving the interatomic forces in a molecule. These forces should be the same relative to the molecule's coordinates, independent of how the molecule is translated or rotated.

The most fundamental aspect of Euclidean neural networks involves categorizing data based on how it transforms under the operations in the Euclidean group E(3), a group in three-dimensional space that contains translations, rotations, and mirroring. These data categories are called irreducible representations (irreps) and are labeled as $\ell = 0, 1, 2, \hdots$ where $\ell=0$ corresponds to a scalar, while $\ell = 1$ corresponds to a three-dimensional vector. Irreps may also possess a parity, which can be either even or odd, indicating whether the representation changes signs when inverted; odd irreps change signs upon inversion, while even irreps remain unchanged. An irreducible representation with $\ell = 1$ and odd parity is termed a vector, representing entities like velocity or displacement vectors. In contrast, an irreducible representation with even parity is referred to as a pseudovector, and it characterizes properties such as angular velocity, angular momentum, and magnetic fields. The input to a Euclidean neural network is concatenated tensors of different data types, for example, a scalar representing a mass concatenated with a vector representing a velocity.

We call a tensor composed of various irreducible representations a \emph{geometric tensor}. In our graph neural network, the equivariant version of vector multiplication involves two geometric tensors and is known as a tensor product $\vx \otimes_{\vw} \vy$. Here, $\vw$ are learnable weights. Our approach employs these tensor products for equivariant message passing, departing from conventional linear operations. For a more in-depth exploration of Euclidean graph neural networks, we refer readers to \citep{e3nn}.

\subsection{Machine learning model}
We construct an equivariant graph self-attention network where the input to the network depends on the chemical structure $\mathcal{G}$ and the atom positions matrix $\mR$ of the specific molecule (see Figure \ref{fig:model_overview}). We exclude hydrogen atoms from the representation of molecules to reduce computational complexity. Every atom/node is represented by a learnable embedding vector $\vx_i$, where the embedding depends on the specific atom type  $Z_i$ (for example, 4 for carbon or 8 for oxygen) and the number of hydrogen atoms connected to that particular atom $N^h_i$. The node/atom input embedding vector is
\begin{equation}
    \vx_i^0 = \left (\text{Emb}(Z_i) ) \ \ || \ \ \text{Emb}(N^h_i)\right ),
\end{equation}
where we denote the concatenation of two vectors with $(\ \cdot\ ||\ \cdot\ )$. We create edges between all atoms in the molecule within a cutoff radius $r_{\text{cut}} = 6 \ \textup{\AA}$. Every edge is represented by a vector of scalars ($\ell = 0$ and even parity) $\mathbf{h}_{ij}^s = \left (  \text{Emb}(E_{ij}) \ || \ d_{ij}\right )$ where $\text{Emb}(E_{ij})$ is an embedding vector depending on the particular bond type $E_{ij}$ (no bond, single bond, or double bond), and $d_{ij} = || \mathbf{r}_i - \mathbf{r}_j||$ is the euclidean distance between the nodes $i$ and $j$. We also construct an embedding of the normalized relative distance between the nodes/atoms, $\hat{\vr}_{ij} = \vr_i -\vr_j$ using spherical harmonics $ Y_m^{\ell}(\hat{\vr}_{ij}/||\hat{\vr}_{ij}||)$, where m is the parity and $\ell$ is the rotation order. 

The layers in the network consist of E(3)-equivariant self-attention/transformer layers \citep{fuchs2020se, liao2023equiformer}, built using the \emph{e3nn} library \citep{e3nn}. For the layers $k = 1, \hdots, K$, we derive messages by deriving queries $\vq^k$, keys $\vk^k$, and value $\vv^k$ as
\begin{align}
    \vq_{i}^k &= \text{Linear}(\vx_i^k)\\
    \vk_{ij}^k &=  \vx_{i}^k\otimes_{\vw^k_{ij, \vk}} Y_m^{\ell}(\hat{\vr}_{ij}/||\hat{\vr}_{ij}||) \\
    \vv_{ij}^k &= \vx_{i}^k\otimes_{\vw^k_{ij, \vv}} Y_m^{\ell}(\hat{\vr}_{ij}/||\hat{\vr}_{ij}||) 
\end{align}
where Linear is a generalization of a regular linear layer for a geometric tensor. The weights of the tensor products $\otimes$ are derived by neural networks, with the invariant edge embeddings as inputs: $ w^k_{ij} = \text{NN}_k(e^s_{ij})$ and $w^v_{ij} = \text{NN}_v(e^s_{ij})$. The self-attention is derived as 
\begin{equation}
    \alpha_{ij}^k = \frac{\exp\left(\vq_i^k\otimes\vk_{ij}^k\right)}{\sum_{j\in\mathcal{N}(i)}\exp\left(\vq_i^k\otimes\vk_{ij}^k\right)}
\end{equation}
where $\vq_i\otimes\vk_{ij}$ maps to a scalar ($\ell = 0$). We aggregate the messages by summing up the weighted messages from all neighboring nodes $\mathcal{N}(i)$
\begin{equation}
    \vx^{k'}_i = \sum_{j\in \mathcal{N}(i)}\alpha_{ij}\vk_{ij}^k.
\end{equation}
In between the self-attention layers, the geometric tensors are updated with equivariant Layer Normalization (LN) \citep{liao2023equiformer} and an equivariant neural network (NN) as
\begin{equation}
    \vx^{k+1}_i = \text{LN}\left(\text{NN}(\vx^{k'}_i)  + \vx^{k}_i\right),  
\end{equation}
where the neural network consists of the generalized linear layers (Linear) and Sigmoid linear units (SiLU) activation functions. The output of the last layer $K$ is an invariant vector $\vx^{K}_i$. Finally, a multilayer perceptron with scalar output is applied.

We train the model by minimizing the mean absolute error, 
\begin{equation}
    \mathcal{L} = \frac{1}{N} \sum_{i=1}^{N} |x_i-\hat{x}_i|, 
\end{equation}
where $N$ is the number of chemical shifts, $x_i$ is the experimentally determined chemical shift, and $\hat{x}_i$ is the predicted one. 

We train the model with multiple conformations and, thereby, multiple graphs for each chemical shift $x_i$. This results in an ensemble of predictions $\hat{x}_i^0, \hdots, \hat{x}_i^j, \hat{x}_i^{N_i}$ for every output $x_i$. We want the mean of this ensemble to be equal to the experimentally determined chemical shift, such that $\frac{1}{N_i} \sum_j \hat{x}_i^j \approx x_i$. Thus, we aim at minimizing $ \left|x_i- \frac{1}{N_i}\sum_{j=1}^{N_i}\hat{x}^j_i \right|$. It follows from the triangle inequality that
\begin{equation}\label{eq:loss}
  \left  |x_i- \frac{1}{N_i}\sum_{j=1}^{N_i}\hat{x}^j_i \right| \leq   \frac{1}{N_i}\sum_{j=1}^{N_i}
  \left  |x_i - \hat{x}^j_i \right|
\end{equation}
hence, we can minimize the right-hand side of the Equation \ref{eq:loss}. This results in the relatively simple conclusion that we, in the training dataset, can add the ensemble of conformations to create a single large training dataset.

\subsection*{Implementation details}
The dimension of the input node embedding $\vx_i^0$ is 128, and the input scalar edge embedding $\ve^0_{ij}$ is 32. The model consists of seven layers where the hidden dimensions between the layers consist of a scalar vector of size 64, 32 tensors with $\ell=1$ and odd parity, and eight tensors with $\ell = 2$ and even parity. Between the self-attention layers, the hidden layer is passed through an equivariant neural network with one hidden layer and a SiLU non-linearity, followed by an equivariant layer normalization. The last layers map the tensors to a scalar vector with 128 dimensions. This vector is passed through a two-layer multilayer perceptron with a hidden dimension of 384 and an output dimension of one.

The batch size of the models is set to 32 except for SG-\mbox{IMP-IR}, where the recommended batch size of 128 is used. The models are optimized using the Adam optimizer \citep{kingma2014adam} starting with a learning rate of 3e-4. We used a small validation set consisting of five percent of the training data for the models trained using only one conformation per molecule. The learning rate decreased during training using the PyTorch ReduceLROnPlateau, which reduces the error when the validation error stops decreasing. A patience of 20 epochs and a reducing factor of 0.1 was used. We did not use a scheduler for the instances when multiple conformations are used. Instead, we trained these models during three epochs, and the learning rate decreases by 0.1 for every new epoch.

The model is implemented using Python 3.9.13, PyTorch version 2.0.0, Cuda version 11.7, PyTorch geometric version 2.3.0, e3nn version 0.5.1, RDKit version 2022.09.5, and GlyLES version 0.5.11. The models are trained using one NVIDIA A100 GPU. The training time per model takes around 30 minutes to an hour.

\subsection{The dataset}\label{sec:datasets}
The dataset consists of $^1$H and $^{13}$C NMR chemical shifts of mono- to trisaccharides. The data is used by CASPER \citep{JANSSON20061003,lundborg2011structural,Dorst2023Carbres} and is based on published data \url{http://www.casper.organ.su.se/casper/liter.php}, including, inter alia, those related to structures of biological interest \citep{Roslund2011CarbRes,Ronnols2013CarbRes,Furevi2022CarbRres}. In detail, it encompasses hydrogen $^1$H and carbon $^{13}$C chemical shifts for 375 carbohydrates in an aqueous solution. Of these are 107 monosaccharides, 153 disaccharides, and 115 trisaccharides. By summing up the individual shifts, the dataset contains 
5356 $^1$H and 4713 $^{13}$C chemical shifts. \emph{GlyLES} \citep{joeres2023glyles} was used to convert the carbohydrates from the IUPAC representation into SMILES representation. The \cite{rdkit} open-source library was used to convert the molecule from the SMILES representation to a graph. RDKit was also used to generate molecular conformations. To obtain 100 conformations per molecule, we generated 200 conformations using the ETKDGv3 method \citep{wang2020improving}. To get a spread in the conformation distribution, we enforced keeping only conformations at a certain distance from each other. By deriving the potential energy using the MMFF94 force field \citep{tosco2014bringing}, we discarded the 100 conformations with the highest energy. 

\section{Acknowledgments}
The simulations were performed on the Luxembourg national supercomputer MeluXina.
The authors gratefully acknowledge the LuxProvide teams for their expert support. This work was supported by grants from the Swedish Research Council (2022-03014) and The Knut and Alice Wallenberg Foundation.
\section{Data availability}
The dataset consisting of $^1$H and  $^{13}$C NMR chemical shifts used in this study will be available on Figshare. Please contact us to access the data before it is submitted. 
\section{Code availability}
The code is available at \url{https://github.com/mariabankestad/GeqShift}.
\section{Author contributions}
M.B. developed and implemented the software, conducted the experiments, and produced the illustrations. J.R., K.D., and G.W. contributed with the data and with expertise in carbohydrates. All four authors M.B, J.R, K.D and G.W took part in writing the manuscript.

\section{Competing interests}
The authors declare no competing interests in this research.

\begin{appendices}



\end{appendices}


\bibliography{bibliography}

\end{document}